\documentclass[letterpaper, 10 pt, conference]{ieeeconf}  

\IEEEoverridecommandlockouts                              

\usepackage{multirow}
\usepackage{graphicx}
\usepackage{subcaption}
\usepackage{amsmath}
\usepackage{amsfonts}
\usepackage{amssymb}

\usepackage{algorithm}
\usepackage{algorithmic}

\usepackage{mathtools}  
\usepackage{bm}         
\usepackage{dsfont}

\newtheorem{definition}{Definition}[section]
\newtheorem{theorem}{Theorem}[section]
\newtheorem{lemma}{Lemma}[section]

\newtheorem{remark}{Remark}[section]
\newtheorem{corollary}{Corollary}[section]

\usepackage{hyperref}   
\usepackage{cleveref}   

\usepackage{url}

\title{\LARGE \bf
Enhancing Certifiable Semantic Robustness\\ via Robust Pruning of Deep Neural Networks
}

\author{Hanjiang Hu$^*$ ~  Bowei Li$^*$  ~ Ziwei Wang$^*$ ~ Tianhao Wei ~ Casidhe Hutchison ~ Eric Sample ~Changliu Liu 
\thanks{*Equal contribution}
\thanks{All authors are with the Robotics Institute, Carnegie Mellon University. Ziwei Wang is also with School of Electrical and Electronic Engineering, Nanyang Technological University, Singapore. The work was done while Ziwei Wang was a postdoc fellow at Carnegie Mellon University.
        {\tt\small hanjianghu@cmu.edu}}%
}

\begin{document}

\maketitle
\thispagestyle{empty}
\pagestyle{empty}

\begin{abstract}
Deep neural networks have been widely adopted in many vision and robotics applications with visual inputs. It is essential to verify its robustness against semantic transformation perturbations, such as brightness and contrast. However, current certified training and robustness certification methods face the challenge of over-parameterization, which hinders the tightness and scalability due to the over-complicated neural networks. To this end, we first analyze stability and variance of layers and neurons against input perturbation, showing that certifiable robustness can be indicated by a fundamental Unbiased and Smooth Neuron metric (USN). Based on USN, we introduce a novel neural network pruning method that removes neurons with low USN and retains those with high USN, thereby preserving model expressiveness without over-parameterization. To further enhance this pruning process, we propose a new Wasserstein distance loss to ensure that pruned neurons are more concentrated across layers. We validate our approach through extensive experiments on the challenging robust keypoint detection task, which involves realistic brightness and contrast perturbations, demonstrating that our method achieves superior robustness certification performance and efficiency compared to baselines.


\end{abstract}

\section{INTRODUCTION}

Deep neural networks (DNNs) have emerged as fundamental components in numerous computer vision and robotics applications, from image classification to pose estimation \cite{he2016deep,kouvaros2023verification,talak2023certifiable}. In safety-critical scenarios such as autonomous driving and human-robot interaction, these DNN-based systems must maintain reliable performance under various environmental conditions, such as seasonal and daylight changes \cite{hu2019retrieval,hu2023seasondepth}, image corruptions and degradations \cite{kong2023robodepth,kong2024robodrive}, and sensor placement and perturbations \cite{hu2022investigating,hu2023robustness}. However, ensuring the robustness of DNNs against such semantic perturbations remains a significant challenge, particularly when formal guarantees are required \cite{luo2025certifying, hu2024pixel,hu2022robustness}.

Current approaches to neural network robustness have primarily focused on adversarial training \cite{madry2018towards} and empirical evaluation methods. While these techniques can improve practical robustness, they fall short of providing the formal verification guarantees essential for safety-critical applications. Certified robustness methods \cite{cohen2019certified,zhang2018efficient} address this limitation by providing mathematical guarantees on model behavior within specified perturbation bounds. However, existing robust training and certification techniques face fundamental scalability challenges due to the over-parameterization of modern deep networks, which leads to loose bounds and computational intractability for larger DNNs \cite{liu2021algorithms}.

\begin{figure}
    \centering
\includegraphics[width=0.95\linewidth]{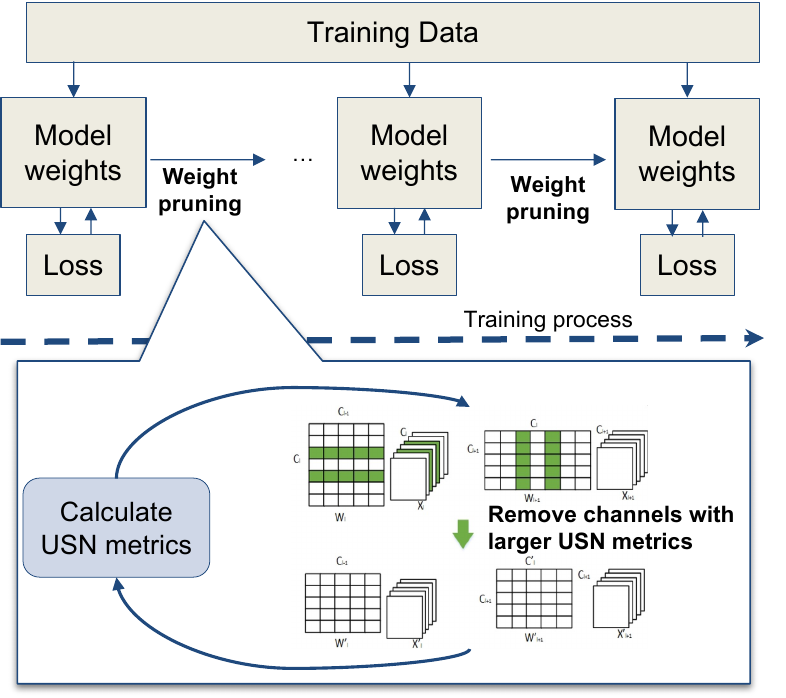}
    \caption{The overview of  model training with the progressive robust pruning pipeline.}
    \label{fig:overview}
\end{figure}

The core issue underlying these challenges is that current robust training and certification methods treat all neurons equally, without considering their individual contributions to model robustness. 
Neurons that exhibit high variance or instability under input perturbations contribute more to the over-approximation errors that fail existing verification algorithms \cite{zhang2018efficient,wei2024improve}. This observation suggests that strategic removal of problematic neurons through structured pruning could simultaneously improve both certification tightness and computational efficiency.

Neural network pruning has been extensively studied mainly for model compression and acceleration \cite{han2016deep,liurethinking,cheng2024survey}, but naive structured pruning may remove important features needed for model expressiveness and robust performance \cite{franklelottery}. Although some existing pruning methods demonstrate a correlation with adversarial robustness \cite{sehwag2020hydra,jordao2021effect,zhangheng2022can}, they do not account for the impact on formal neural network verification based on neuron statistics. 
The challenge lies in theoretically identifying unstable neurons that are truly detrimental to robustness certification while preserving the model's expressive capacity for the underlying task. 

To this end, we propose a novel approach that integrates robustness-aware pruning with formal certification guarantees through the introduction of Unbiased and Smooth Neuron (USN) metrics, which quantify the bias and variance characteristics of individual neurons under semantic perturbations. We highlight that the proposed USN metric is a generalized form of the signal-to-noise ratio metrics \cite{wei2024improve}, which additionally bridges the metrics with statistical principles without normalization, focusing on complex semantic perturbation rather than simple $\ell_\infty$-bounded perturbations \cite{wei2024improve}.
This provides a principled criterion for identifying neurons that contribute most to certification looseness. We develop a progressive training and pruning pipeline that simultaneously optimizes task performance and USN metrics, as shown in \Cref{fig:overview}. Wasserstein distance regularization is further adopted to encourage concentrated pruning patterns while preserving essential representational capacity. As one of the representative regression tasks, we evaluate the proposed method on the challenging keypoint detection task under image-based realistic semantic perturbations, demonstrating that USN-guided pruning consistently outperforms both non-pruned models and random pruning baselines across multiple base architectures and perturbation magnitudes.
The main contributions of this work are as follows:
\begin{itemize}
\item We establish a theoretical connection between neuron-level statistics and probabilistic robustness certification bounds, providing the foundation for robustness-aware pruning.
\item We introduce the Unbiased and Smooth Neuron (USN) metrics that quantify individual neuron contributions to certification tightness under semantic perturbations.
\item We propose a progressive training pipeline that integrates USN-guided pruning with Wasserstein distance regularization to achieve concentrated, structure-preserving pruning patterns.
\item We demonstrate superior certification performance on keypoint detection tasks against brightness and contrast perturbations, showing that the proposed pruning can simultaneously improve robustness and computational efficiency.
\end{itemize}








\section{PROBLEM FORMULATION}
In this section, we first formally formulate the key point detection neural networks as feedforward layers and non-linear ReLU activation layers. Then, we define the certifiable robustness within the local semantic perturbation set.
\subsection{Neural Networks with Neuron Characterization}
Modern certifiable image-based deep neural network models are typically based on ResNet \cite{he2016deep,kouvaros2023verification,talak2023certifiable,luo2025certifying}, including convolutional neural networks (CNNs) and linear layers with nonlinear activation functions. Since convolution can be seen as sparse matrix multiplication with shared weights of convolutional kernels, the fundamental components of these neural networks for regression tasks (e.g. keypoint detection) are linear layers with nonlinear activation functions, as defined below.
\begin{definition}[Deep Neural Networks for Regression]
\label{def:nn}
Let $f^L: \mathbb{R}^{d_0} \rightarrow \mathbb{R}^{d_L}$ be an $L$-layer feedforward layers with nonlinear activation layers defined as:
\begin{equation}
\label{eq:nn}
f^L(x) = g^L \circ  \sigma^{L-1} \circ g^{L-1} \circ \sigma^{L-2} \circ \cdots \circ \sigma^1 \circ g^1(x),
\end{equation}
where each linear layer $g^i: \mathbb{R}^{d_{i-1}} \rightarrow \mathbb{R}^{d_i}$ is given by $g^i(z^{i-1}) = W^i z^{i-1} + b^i$ with weight matrix $W^i \in \mathbb{R}^{d_i \times d_{i-1}}$ and bias vector $b^i \in \mathbb{R}^{d_i}$, and activation layer is defined as  $z^i = \sigma^i(g^i(z^{i-1}))$ with nonlinear activation function $\sigma^i: \mathbb{R}^{d_i} \rightarrow \mathbb{R}^{d_i}$, $i=1,2,\dots,L$ and $z^0=x$.
\end{definition}

Furthermore, we define the output of each neuron on the layer before each activation layer as follows, as the nonlinearity of the neural network in \Cref{eq:nn} highly depends on the output of pre-activation neurons.
\begin{definition}[Pre-activation Neuron and Layer Output]
\label{def:neuron_output}
Denote the $j$-th neuron output after the linear layer $f^i$  as $f^i_j(x) = w^i_j \cdot z^{i-1} + b^i_j \in \mathbb{R}$, $j=0,1,\dots,d_i-1$ where $w^i_j$ is the $j$-th row of $W^i$, $z^{i-1} = \sigma(f^{i-1}(x))$ and the layer output $f^i(x)$ is the aggregated vector of $f^i_j(x)$ as  $f^i(x)=[f^i_0(x), f^i_1(x), \dots, f^i_{d_i-1}(x)]^T \in \mathbb{R}^{d_i}$.
\end{definition}

\subsection{Semantic Perturbation and Certifiable Robustness}
Based on the neural networks in \Cref{def:nn} with each neuron output in \Cref{def:neuron_output}, we can define the certifiable robustness against semantic perturbation. We first define the semantic perturbation set as follows.
\begin{definition}[Semantic Perturbation Set]
\label{def:semantic_pert}
Given an input $x_0 =h(s_0) \in \mathbb{R}^{d_0}$ with a continuous semantic transformation  function $h(\cdot): \mathbb{R}^s \rightarrow \mathbb{R}^{d_0}$, define the bounded perturbation set as $B_p^h(x_0,\epsilon) = \{h(s) \in \mathbb{R}^{d_0} : \|s - s_0\|_p \leq \epsilon\}$, where the norm $\|\cdot\|_p$ is defined in semantic perturbation space $\mathbb{R}^s$  with perturbation radius $\epsilon > 0$.
\end{definition}
\begin{remark}
    Since we are dealing with keypoint detection as a regression task using neural networks under semantic perturbation on the input image, e.g., brightness and contrast, the perturbation space $\mathbb{R}^s$ is usually the 1D real space $s=1$, and the assumption holds that semantic transformation function is continuous. 
\end{remark}

Based on the semantic perturbation over the input of the neural network, we then define the robustness certification problem as follows. 

\begin{definition}[Robustness Certification]
\label{def:robustness_certification}
Given the certification criteria with radius $\delta$ under $\|\cdot\|_q$ norm in the output space $\mathbb{R}^{d_L}$ of neural networks $f^L$ in \Cref{def:nn}, the robustness certification problem is to verify whether the following condition holds for semantic perturbation set $B_p^h(x_0,\epsilon)$ in \Cref{def:semantic_pert},
\begin{align}
\label{eq:cert_goal}
    \forall x \in B_p^h(x_0,\epsilon), \|f^L(x) - f^L(x_0)\|_q \leq \delta.
\end{align}
\end{definition}
\begin{remark}
    In the case of the keypoint detection task with deep neural networks under semantic perturbation on the input image, the original output is the heat map. However, with the addition of an extra layer of differentiable spatial-to-numerical transformation \cite{nibali2018numerical}, the entire neural network aligns well with \Cref{def:nn} for the keypoint regression task.
    The certification criteria is defined as the maximal pixel deviation of all keypoints away from those of $f^L(x_0)$, i.e., $\delta$ in pixels under $\ell_\infty$ norm for $d_L$ keypoints.
\end{remark}

\section{METHODOLOGY}
Equipped with the definitions of neural networks and pre-activation neuron outputs and the goal of robustness certification under semantic perturbation, in this section, we first analyze the mean and variance of neuron output distribution given the semantic perturbation. Then we introduce the unbiased and smooth neuron metric as an empirical estimate from Monte Carlo sampling of the semantic perturbation set, based on which we introduce a neural network pruning training pipeline to gain more stable and low-variance neurons. Furthermore, a regularization term is proposed based on the Wasserstein distance for concentrated pruning. 

\subsection{Neuron Stability and Variance Analysis}
Given input perturbation of neural networks, the robustness certification in \Cref{def:semantic_pert} is determined by the nonlinearity and Lipschitz continuity of neural networks \cite{virmaux2018lipschitz, pauli2021training,wei2024improve}. We present the following Lemma to quantify the Lipschitz bound propagation from intermediate layers to neural network output.

\begin{lemma}[Layer-to-Output Lipschitz Bound]
\label{lem:lipschitz_bound}
For a neural network $f^L$ as defined in \Cref{def:nn} and original input $x_0$, the output deviation of perturbed input $x\in B_p^h(x_0,\epsilon)$ under norm $\|\cdot\|_q, q \geq2,$  can be bounded in terms of any intermediate layer $i=1,\dots,L-1$ as,
\begin{align}
    \|f^L(x) - f^L(x_0)\|_q \leq  C_i \|f^i(x) - f^i(x_0)\|_2,
\end{align}
where the constant $C_i = \|W^i\|_2\prod_{k=i+1}^{L} \|W^k\|_2 \cdot L_{\sigma^{k-1}}$ $L_{\sigma^k}$ and the $\ell_2$ Lipschitz constant of activation function $\sigma^k$.
\end{lemma}

\begin{proof}
By the induced norm of $i$-th layer weight $W^i$ and Lipschitz constant of activation function $\sigma^i$, we have
\begin{align}
&\|f^{L}(x) - f^{L}(x_0)\|_q \leq \|f^{L}(x) - f^{L}(x_0)\|_2 \\
\leq& \|W^{L}\|_2 \cdot L_{\sigma^{L-1}} \cdot \|f^{L-1}(x) - f^{L-1}(x_0)\|_2 \leq \dots \\ \leq & \left(\|W^i\|_2\prod_{k=i+1}^{L} \|W^k\|_2  L_{\sigma^{k-1}}\right) \|f^i(x) - f^i(x_0)\|_2.
\end{align}
Therefore, the bound holds for any layer $i, i\leq L-1$.
\end{proof}
\Cref{lem:lipschitz_bound} bridges the certification goal of \Cref{eq:cert_goal} with layer-wise deviation given input perturbation, which are highly related to the stability and variance of the intermediate pre-activation neuron outputs in \Cref{def:neuron_output}. 
Unlike \cite{wei2024improve}, we investigate the stability of neurons based on the statistical distribution of neuron outputs at each layer from semantic perturbation sampling.

\begin{definition}[Neuron Output Distribution]
\label{def:distribution_neuron}
Given input sample $x_0\sim p(x_0)$, suppose the perturbed input $p(x\mid x_0)$ is uniformly sampled from $B_p^h(x_0,\epsilon)$, i.e., $x\mid x_0\sim\mathcal{U}(B_p^h(x_0,\epsilon))$, the $j$-th pre-activation neuron output on layer~$i$  $f^i_j(x)$ has the mean of $\mathbb{E}_{x\sim p(x\mid x_0)}f^i_j(x)$ and variance of $\text{Var}_{x\sim p(x\mid x_0)}f^i_j(x)$.
\end{definition}

Based on the neuron output distribution under perturbation sampling, we have the following Theorem showing how the mean and variance of the neuron output distribution affect the robustness certification goal in \Cref{eq:cert_goal}.


\begin{theorem}[Probabilistic Robustness Certification]
\label{thm:robustness_bound}
For any sample $x_0\sim p(x_0)$ and the uniformly perturbed sample $x\mid x_0\sim\mathcal{U}(B_p^h(x_0,\epsilon))$, the robustness certification goal $\|f^L(x) - f^L(x_0)\|_q \leq \delta$ in \Cref{eq:cert_goal} holds with confidence of $1-\alpha$ if the following inequalities hold for each neuron $j=1,2,\dots,d_i$ on layer $i=1,2,\dots,L-1$
\begin{align}
\left|\mathbb{E}_{x\sim p(x\mid x_0)}f^i_j(x) - f^i_j(x_0)\right| &\leq \frac{\delta}{2C_i\sqrt{d_i}},
\label{eq:bias_bound}\\
\text{Var}_{x\sim p(x\mid x_0)}f^i_j(x) &\leq \frac{\alpha\delta^2}{4C_i^2 d_i^2(L-i)},
\label{eq:variance_bound}
\end{align}
where $C_i$ is Lipschitz bound from \Cref{lem:lipschitz_bound}.
\end{theorem}

\begin{proof}
For the random variable of neuron output $f^i_j(x)$ from \Cref{def:distribution_neuron}, first apply Chebyshev's inequality and variance bound in \Cref{eq:variance_bound} as follows,
\begin{align*}
    P(| f^i_j(x) - \mathbb{E}f^i_j(x)|\geq \frac{\delta}{2C_i\sqrt{d_i}})\leq \frac{\text{Var}f^i_j(x)}{(\frac{\delta}{2C_i\sqrt{d_i}})^2} 
    \leq \frac{\alpha}{d_i(L-i)}.
\end{align*}
Therefore, with a probability of at least $1-\frac{\alpha}{d_i(L-i)}$, $|f^i_j(x) - \mathbb{E}_{x\sim p(x|x_0)}f^i_j(x)| \leq \frac{\delta}{2C_i\sqrt{d_i}}$ holds. Then by triangle inequality and the bias bound in \Cref{eq:bias_bound}, we have the following hold with probability of at least $1-\frac{\alpha}{d_i(L-i)}$,
\begin{align}
    &|f^i_j(x) - f^i_j(x_0)|\leq |f^i_j(x) - \mathbb{E}_{x\sim p(x|x_0)}f^i_j(x)| \notag\\&+ |\mathbb{E}_{x\sim p(x|x_0)}f^i_j(x) - f^i_j(x_0)| \leq \frac{\delta}{C_i\sqrt{d_i}} 
    \label{eq:dev_ij}.
\end{align}
By union bound of \Cref{eq:dev_ij} along each neuron $j$ along layer $i$, the $\ell_2$ norm of layer deviation $f^i(x) - f^i(x_0)$ can be upper bounded below with probability of at least $1-\frac{\alpha}{(L-i)}$,
\begin{align}
    \|f^i(x) - f^i(x_0)\|_2 = \sqrt{\sum_{j=1}^{d_i}|f^i_j(x) - f^i_j(x_0)|^2}\leq \frac{\delta}{C_i}.
    \label{eq:dev_i}
\end{align}
Again, based on \Cref{lem:lipschitz_bound}, by union bound of \Cref{eq:dev_i} along layers from $i$ to $L-1$, we have $\|f^L(x) - f^L(x_0)\|_q \leq C_i \|f^i(x) - f^i(x_0)\|_2\leq \delta$ hold with  probability of at least $1-
\alpha$, which concludes the proof.
\end{proof}
\begin{remark}
    We remark that the probabilistic certification will naturally become the deterministic robustness certification in \Cref{def:robustness_certification} by letting $\alpha\rightarrow0$. In practice, we usually want the strongest robustness certification with $\delta=0$. Therefore, we need to make the upper bounds of \Cref{eq:bias_bound} and \Cref{eq:variance_bound} close to 0 during model training.
    \end{remark}


\subsection{Unbiased and Smooth Neuron (USN) Metrics}
During empirical model training, we need to estimate the statistics of the neuron output distribution in \Cref{def:distribution_neuron} through finite Monte Carlo sampling. Therefore, in this section, we define the unbiased and smooth neuron metrics to  empirically estimate the statistics in \Cref{eq:bias_bound} and \Cref{eq:variance_bound}, respectively.



\begin{definition}[Unbiased and Smooth Neuron Metrics]
\label{def:usn}
Given the input $x_0$, the unbiased and smooth metrics for layer $i$ are defined as:
\begin{align}
\mathcal{L}^i_{\text{unbiased}} := & \frac{1}{m} \sum_{k=1}^m \|f^i(x_k) - f^i(x_0)\|_1,
\label{eq:unbiased}
\\
\mathcal{L}^i_{\text{smooth}} :=& \frac{1}{m} \sum_{k=1}^m \|f^i(x_k) - f^i(x_0)\|_2^2,
\label{eq:smooth}
\end{align}
where $m$ perturbed input are uniformly sampled from $x_k\in B_p^h(x_0,\epsilon), k=1,2,\dots,m$.
\end{definition}

Even though \Cref{eq:unbiased} and \Cref{eq:smooth} are focusing on $\ell_1$ and $\ell_2$ norm of the difference between original layer output $f^i(x_0)$ and perturbed layer output $f^i(x_k)$, they have fundamental nuance in the stability and variance of the neuron output in the sense of \Cref{eq:bias_bound} and \Cref{eq:variance_bound}, respectively. We present the following lemma to formally characterize these relationships.

\begin{lemma}[USN Metrics and Robustness Bounds]
\label{lem:usn_bounds}
For the unbiased and smooth neuron metrics defined in \Cref{eq:unbiased} and \Cref{eq:smooth} with sufficient samples, the following relationships with \Cref{eq:bias_bound} and \Cref{eq:variance_bound} hold:
\begin{align}
\mathcal{L}^i_{\text{unbiased}} &= \sum_{j=1}^{d_i} |\mathbb{E}_{x\sim p(x|x_0)}f^i_j(x) - f^i_j(x_0)|,\label{eq:unbiased_relation}\\
\mathcal{L}^i_{\text{smooth}} &= \sum_{j=1}^{d_i} [\text{Var}_{x}f^i_j(x)+ |\mathbb{E}_{x}f^i_j(x) - f^i_j(x_0)|^2].
\label{eq:smooth_relation}
\end{align}
\end{lemma}

\begin{proof}
For the first relation in \Cref{eq:unbiased_relation}, by definition of the $\ell_1$ norm and law of large number:
$\mathcal{L}^i_{\text{unbiased}} 
= \frac{1}{m} \sum_{k=1}^m \sum_{j=1}^{d_i} |f^i_j(x_k) - f^i_j(x_0)| 
= \sum_{j=1}^{d_i}\mathbb{E}_{x\sim p(x|x_0)}|f^i_j(x) - f^i_j(x_0)|$.
Since $x_0$ is deterministic given the conditioning, we have $|f^i_j(x) - f^i_j(x_0)| = |\mathbb{E}_{x\sim p(x|x_0)}f^i_j(x) - f^i_j(x_0)|$ in expectation, establishing \Cref{eq:unbiased_relation}.
Similarly, for the second relation in \Cref{eq:smooth_relation}, based on bias-variance decomposition, we have
\begin{align*}
\mathcal{L}^i_{\text{smooth}} 
&= \sum_{j=1}^{d_i} \mathbb{E}_{x}[f^i_j(x) - \mathbb{E}(f^i_j(x)) + \mathbb{E}(f^i_j(x)) - f^i_j(x_0)]^2 \\
&= \sum_{j=1}^{d_i} [\text{Var}_{x}f^i_j(x) +(\mathbb{E}_{x}f^i_j(x) - f^i_j(x_0))^2],
\end{align*}
which concludes the proof.
\end{proof}
Combining \Cref{lem:usn_bounds} and \Cref{thm:robustness_bound}, we present the following robustness certification theorem with the USN metric conditions.
\begin{corollary}[USN Necessary Bound Conditions]
\label{thm:usn_cert}
If \Cref{eq:bias_bound} and \Cref{eq:variance_bound} hold for any neuron $i$ on layer $j$ in \Cref{thm:robustness_bound}, we have the following upper bounds for $\mathcal{L}^i_{\text{unbiased}}$ and $\mathcal{L}^i_{\text{smooth}}$,
\begin{align}
\mathcal{L}^i_{\text{unbiased}} \leq \frac{\delta\sqrt{d_i}}{2C_i},~
\mathcal{L}^i_{\text{smooth}}\leq \frac{\delta^2}{4C_i^2}\left(\frac{\alpha}{d_i(L-i)} + 1\right).
\label{eq:usn_bound}
\end{align}
\end{corollary}

\begin{proof}
The upper bounds can be obtained by applying \Cref{eq:bias_bound} and \Cref{eq:variance_bound} to  \Cref{eq:unbiased_relation} and \Cref{eq:smooth_relation}.  
\end{proof}
\begin{remark}
Even though \Cref{eq:usn_bound} is not a sufficient condition for robustness certification for general $\delta>0, \alpha>0$, but it is aligned with the robustness certification goal in \Cref{eq:cert_goal} when $\delta\rightarrow0, \alpha\rightarrow0$, showing that minimizing $\mathcal{L}^i_{\text{unbiased}}$ and $\mathcal{L}^i_{\text{smooth}}$ will lead to certification goal.
    We remark that the signal-to-noise ratio (SNR) losses in \cite{wei2024improve} are special cases of unbiased and smooth neuron metrics, where they are normalized by $\|f^i(x_0)\|$. When $\|f^i(x_0)\|$ significantly increases during model training, even if SNR losses can be greatly reduced, the upper bounds of \Cref{eq:bias_bound,eq:variance_bound} do not necessarily hold, and therefore the certification goal would fail when $\delta\rightarrow0, \alpha\rightarrow0$.
    Besides, ours can handle more general semantic perturbations while SNR losses mainly focus on $\ell_\infty$-bounded perturbations.
\end{remark}

\subsection{Wasserstein Distance for USN Regularization}
Since smaller USN metrics of $\mathcal{L}^i_{\text{smooth}}$ and $\mathcal{L}^i_{\text{smooth}}$ in \Cref{def:usn} can inherently ensure the robustness of the neural network, we adopt structured pruning \cite{zhangheng2022can,chen2022linearity} of the neurons with larger USN metrics.
To ensure concentrated and coherent pruning patterns across network layers, we introduce a Wasserstein distance regularization that promotes structured sparsity. The Wasserstein distance \cite{panaretos2019statistical}, also known as the Earth Mover's Distance, measures the minimum cost to transform one probability distribution into another.

\begin{definition}[Wasserstein Distance]
\label{def:wasserstein}
For two discrete probability distributions $\mu = \sum_{i=1}^n a_i \delta_{x_i}$ and $\nu = \sum_{j=1}^m b_j \delta_{y_j}$ with $\sum_i a_i = \sum_j b_j = 1$, the 2-Wasserstein distance is defined as:
\begin{equation}
\mathcal{W}_2(\mu, \nu) = \min_{\pi \in \Pi(\mu, \nu)} \left(\sum_{i,j} \pi_{ij} \|x_i - y_j\|_2^2\right)^{1/2},
\end{equation}
where $\Pi(\mu, \nu)$ represents the set of all joint distributions with marginals $\mu$ and $\nu$, and $\pi_{ij}$ denotes the transport plan indicating how much probability mass moves from $x_i$ to $y_j$.
\end{definition}

In the context of neural network pruning, we apply the Wasserstein distance to align the importance distributions of neurons across layers. Drawing from the USN relations in \Cref{lem:usn_bounds}, we define the importance score for each neuron based on its contribution to both the unbiased and smooth metrics. For each layer $i$, the neuron-wise contribution to the USN metrics can be decomposed as:
\begin{align}
\mathcal{L}^i_{\text{unbiased},j} &= |\mathbb{E}_{x\sim p(x|x_0)}f^i_j(x) - f^i_j(x_0)|,
\label{eq:neuron_unbiased}\\
\mathcal{L}^i_{\text{smooth},j} &= \text{Var}_{x}f^i_j(x)+ |\mathbb{E}_{x}f^i_j(x) - f^i_j(x_0)|^2. \label{eq:neuron_smooth}
\end{align}
Therefore, the importance score $\mathcal{A}^i_j$ for neuron $j$ in layer $i$ is then defined as 
the following dimensionless ratio between the smooth metric (with dimension of squared $\ell_2$ norm in \Cref{eq:smooth}) and the square of the unbiased metric (with dimension of $\ell_1$ norm in \Cref{eq:unbiased}),
\begin{equation}
\label{eq:importance}
\mathcal{A}^i_j = \frac{\mathcal{L}^i_{\text{smooth},j}}{({\mathcal{L}^i_{\text{unbiased},j}}^2 + \epsilon_{usn})\cdot d_i},
\end{equation}
where 
where $\mathcal{L}^i_{\text{smooth},j}$ and $\mathcal{L}^i_{\text{unbiased},j}$ are the neuron-wise contributions in \Cref{eq:neuron_unbiased} and \Cref{eq:neuron_smooth}, $\epsilon_{usn}$ is a small regularization constant to prevent division by zero, and $d_i$ provides layer-wise normalization to ensure dimensional consistency and comparability across layers of different widths.
Note that in practice, we conduct structured pruning by calculating the importance score $\mathcal{A}$ for each coarser-grained channel of neurons \cite{franklelottery,zhangheng2022can,cheng2024survey}, which identifies the channels with the most unstable neurons for pruning. 



To encourage concentrated pruning, we define the target distribution $\mathcal{E}^i$ using percentile-based thresholding:
\begin{equation}
\mathcal{E}^i_j = \begin{cases}
\frac{1}{d_i}, & \text{if } \mathcal{A}^i_j > \text{percentile}(\mathcal{A}^i, (1-\rho) \times 100) \\
0, & \text{otherwise}
\end{cases}
\end{equation}
where $\rho$ is the target pruning ratio.
Based on \Cref{def:wasserstein}, the Wasserstein regularization loss for layer $i$ is then:
\begin{equation}
\label{eq:W-loss}
\mathcal{L}_{\text{W}}^i = \mathcal{W}_2(\mathcal{M}^i, \mathcal{E}^i).
\end{equation}
This regularization encourages the network to develop clear distinctions between important and unimportant neurons based on their USN characteristics, facilitating more effective structured pruning by promoting concentrated removal of the most unstable neurons.

\subsection{Integrated Training and Pruning Pipeline}
We finally propose a progressive training pipeline that integrates robust training with dynamic pruning based on USN metrics. The training process consists of multiple phases where the pruning ratio gradually increases, allowing the network to adapt to the reduced capacity.

The total loss function combines multiple components:
\begin{equation}
\mathcal{L}_{\text{total}} = \mathcal{L}_{\text{task}} + \sum_{i \in \mathcal{I}_{\text{prune}}} \left(\lambda_u\mathcal{L}^i_{\text{unbiased}} + \lambda_s\mathcal{L}^i_{\text{smooth}} + \lambda_W \mathcal{L}_{\text{W}}^i\right), \notag
\end{equation}
where task loss $\mathcal{L}_{\text{task}}$ comes from the task-specific domain (e.g. keypoint detection literature \cite{luo2025certifying}) and $\mathcal{I}_{\text{prune}}$ is the set of layers to be pruned. 
The progressive pruning schedule is defined as:
\begin{equation}
\label{eq:pr_sch}
\rho(t) = \begin{cases}
0, & \text{if } t < t_{\text{start}} \\
\frac{\rho}{N_{\text{steps}}} \left\lfloor \frac{t - t_{\text{start}}}{t_{\text{interval}}} \right\rfloor, & \text{if } t_{\text{start}} \leq t \leq t_{\text{end}} \\
\rho, & \text{if } t > t_{\text{end}}
\end{cases}
\end{equation}
where $t$ is the current epoch, $\rho$ is the final target pruning ratio, $N_{\text{steps}}$ is the number of pruning steps, and $t_{\text{interval}}$ is the interval between pruning operations.

\begin{algorithm}
\caption{Progressive USN-Guided Training and Pruning}
\label{alg:progressive_training}
\begin{algorithmic}[1]
\REQUIRE Neural network $f_\theta^L$, training data $\mathcal{D}$, semantic perturbation set $B_p^h(\cdot, \epsilon)$, final pruning ratio $\rho$, pruning steps $N_{\text{steps}}$, learning rate $\alpha$, pruning layer set $\mathcal{I}_{\text{prune}}$
\ENSURE Trained and progressively pruned network
\STATE Initialize: $\rho_{\text{current}} \leftarrow 0$, importance $\{\mathcal{A}^i \leftarrow \mathbf{0}\}_{i=1}^{L-1}$
\FOR{epoch $t = 1$ to $T$}
    \STATE Update pruning ratio: $\rho_{\text{current}} \leftarrow \rho(t)$ according to schedule in \Cref{eq:pr_sch}
    \FOR{each batch $\{x_b, y_b\}$ in $\mathcal{D}$}
        \STATE Sample semantic perturbations $x\leftarrow B_p^h(x_b,\epsilon)$
        \STATE Compute task loss: $\mathcal{L}_{\text{task}}$
        \FOR{each layer $i \in \mathcal{I}_{\text{prune}}$}
            \STATE Compute USN metrics $\mathcal{L}^i_{\text{unbiased}}, \mathcal{L}^i_{\text{smooth}}$ based on \Cref{eq:unbiased} and \Cref{eq:smooth} 
            \STATE Compute neuron importance $\mathcal{A}^i_j$ based on \Cref{eq:importance}
            \STATE Compute Wasserstein loss $\mathcal{L}_{\text{W}}^i$ based on \Cref{eq:W-loss}
        \ENDFOR
        \STATE Total loss: $\mathcal{L}_{\text{total}} \leftarrow \mathcal{L}_{\text{task}} + \sum_{i \in \mathcal{I}_{\text{prune}}} (\lambda_u\mathcal{L}^i_{\text{unbiased}} + \lambda_s\mathcal{L}^i_{\text{smooth}} + \lambda_W \mathcal{L}_{\text{W}}^i) $
        \STATE Backpropagate and update parameters: $\theta \leftarrow \theta - \alpha \nabla_\theta \mathcal{L}_{\text{total}}$
    \ENDFOR
    \IF{$t$ is pruning epoch}
        \FOR{each layer $i \in \mathcal{I}_{\text{prune}}$}
            \STATE Compute pruning threshold: $\tau^i \leftarrow \text{percentile}(\mathcal{A}^i, (1-\rho_{\text{current}}) \times 100)$
            \STATE Generate pruning mask: $\mathcal{P}^i_j \leftarrow \mathds{1}[\mathcal{A}^i_j \geq \tau^i]$
            \STATE Apply structured pruning by keeping masked neurons: $\theta \leftarrow \theta(\mathds{1}[\mathcal{P}^i_j = 1])$
        \ENDFOR
        \STATE Reset  importance: ${\mathcal{A}^i \leftarrow \mathbf{0}}_{i=1}^{L-1}$
    \ENDIF
\ENDFOR
\RETURN Progressively pruned network  $\theta$
\end{algorithmic}
\end{algorithm}

This progressive approach allows the network to gradually adapt to reduced capacity while maintaining performance. The USN metrics guide the pruning process by identifying neurons that exhibit high variance or bias under semantic perturbations, ensuring that the most stable and reliable neurons are preserved. The Wasserstein regularization promotes coherent pruning patterns that maintain the network's structural integrity and representational power.


\section{EXPERIMENTS}
In this section, we evaluate our USN-guided pruning approach by answering two key research questions: 1) \textit{Does USN-guided pruning improve robustness certification performance compared to random pruning and unpruned baselines under realistic semantic perturbations?} 2) \textit{What are the effects of pruning ratio and Wasserstein regularization on the trade-off between certification accuracy, model expressiveness, and verification efficiency?} The answer to the first question will be found in \Cref{sec:results_comparison} through comparisons on CNN7 and ResNet18 architectures, while the second question is addressed in \Cref{sec:ablation} through systematic ablation studies. Prior to those, we first introduce the experimental setup with datasets, training procedures, and evaluation metrics in \Cref{sec:exp_setup}.


\subsection{Experimental Setup}
\label{sec:exp_setup}

\paragraph{Training and Pruning Details} 
Following the literature of keypoint detection \cite{luo2025certifying}, we use the same aircraft dataset with 24 keypoints per image and split them into train/val/test with fixed seeds. We adopt backbones of different architectures: a  seven-layer CNN (CNN7) from \cite{carlini2017towards,zhang2019theoretically} and ResNet18 \cite{he2016deep}.
Models are trained for 200 epochs with Adam (learning rate $\alpha=0.01$), batch size of $64$, and task losses from \cite{luo2025certifying}. We inject small photometric perturbations (brightness $\pm\frac{1}{255}$, contrast $\pm0.01$), apply USN regularization on the convolution layers, and perform progressive channel-level pruning to $\rho=0.1, 0.2$ over $N_{\text{steps}}=200$ steps in \Cref{alg:progressive_training}. Note that channel-level pruning uses the same technique as neuron-wise pruning but with different granularity, and is much more efficient in the literature of structured pruning \cite{franklelottery,zhangheng2022can,cheng2024survey}. We also conduct an ablation of the Wasserstein regularization with $\lambda_W\!\in\!\{0,10\}$. During the model training, we monitor the keypoint task loss on clean data and uniformly perturbed ones and the best checkpoint is chosen by the lowest task loss on the validation set.  Experiments run on a workstation with four NVIDIA RTX A6000 GPUs. 

\paragraph{Certification  Metrics and Baselines} We certify all the models over brightness shifts $\{\pm2,\pm5\}$ pixels ($\epsilon\!\in\!\{2/255,5/255\}$ for brightness transformation $h$ in \Cref{def:semantic_pert}) and contrast scalings $\{\pm0.01,\pm0.02,\pm0.05\}$ ($\epsilon\!\in\!\{0.01,0.02,0.05\}$ for contrast transformation $h$ in \Cref{def:semantic_pert}). For each test image, we crop  and resize to $64{\times}64$ with the output of the 24 keypoints of this frame, and check whether all predicted keypoints stay within the pixel error bound of $1$\,px. That being said, the robustness certification problem is with $d_L=24, \delta=1, q=\infty$ in \Cref{def:robustness_certification}. We adopt the solver \texttt{ModelVerification.jl} \cite{wei2025modelverification} to return \textit{Holds}, \textit{Violated}, or \textit{Unknown} for the certification goal in \Cref{eq:cert_goal}. We adopt the metric of verification accuracy, which is computed as the proportion of test images with certification goal achieved. We also compare the verification time for efficiency and the number of correctly predicted and verified keypoints for fidelity as fine-grained metrics. The baselines are the ones without pruning and with random pruning under multiple pruning rates.







\begin{figure}
    \centering
    \includegraphics[width=0.9\linewidth]{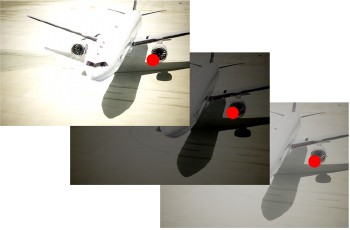}
    \caption{Samples of keypoint detection under changing semantic perturbation of images.}
    \label{fig:image_example}
\end{figure}

\begin{figure}
    \centering
    \includegraphics[width=0.95\linewidth]{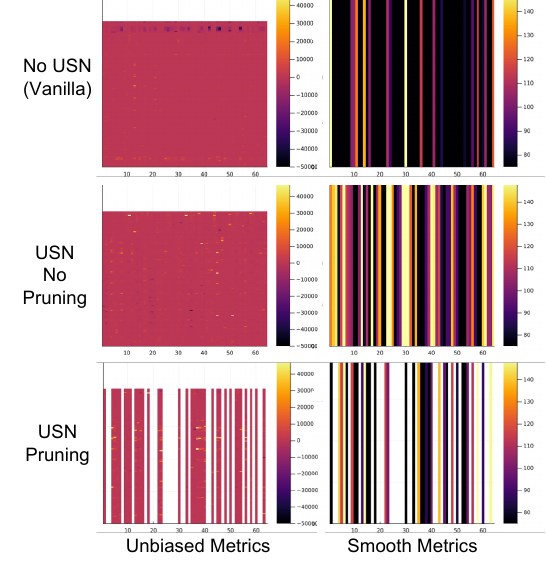}
    \caption{Visualization of USN metrics (color-coded) w.r.t. flattened channels of neurons (each column represents one channel and the horizontal axes show different channels) after vanilla training without USN, robust training with USN but without pruning, and robust training and pruning with USN (ours). We can see that our robust pruning can significantly reduce the neurons with high unbiased and smooth metrics compared to robust training with USN without pruning.}
    \label{fig:visualization}
\end{figure}

\subsection{Results Comparison}
\label{sec:results_comparison}

\paragraph{CNN7}
Table~\ref{tab:r_pr} shows comparison of \emph{USN-guided pruning} with \emph{random pruning} at matched rates under brightness ($\pm2$, $\pm5$) and contrast ($\pm0.01$, $\pm0.02$, $\pm0.05$) perturbations.
USN guidance consistently matches or exceeds random pruning across all magnitudes, with the largest gains appearing at the highest contrast; under stronger brightness shifts, USN also improves over the random pruning baseline.
Overall, USN pruning at a rate $\rho=0.2$ provides the most consistent accuracy across perturbations, indicating that channels selected by USN is more robust than random pruning due to the removal of unstable and high-variance neurons.

\begin{table}[htbp]
    \centering
    \caption{CNN7: USN-guided vs. random pruning under different pruning ratio $\rho$ against brightness/contrast perturbations}
    \label{tab:r_pr}
    \setlength{\tabcolsep}{4pt}
    \renewcommand{\arraystretch}{1.30}
    \begin{tabular}{|c|c|c|c|c|c|c|}
        \hline
        \multirow{2}{*}{$\rho$} & \multirow{2}{*}{Pruning Rule} &
        \multicolumn{2}{c|}{Brightness} & \multicolumn{3}{c|}{Contrast} \\
        \cline{3-7}
        & & $\pm2$ & $\pm5$ & $\pm0.01$ & $\pm0.02$ & $\pm0.05$ \\
        \hline
        \multirow{2}{*}{0.1} & Random & 0.9545 & 0.9025 & \textbf{0.9805} & 0.9350 & 0.8051 \\ 
        \cline{2-7}
        & USN-guided        & \textbf{0.9740} & \textbf{0.9285} & \textbf{0.9805} & \textbf{0.9545} & \textbf{0.8961} \\ 
        \hline
        \multirow{2}{*}{0.2} & Random & 0.9805 & 0.9155 & 0.9805 & 0.9610 & 0.9285 \\ 
        \cline{2-7}
        & USN-guided         & \textbf{0.9870} & \textbf{0.9480} & \textbf{0.9935} & \textbf{0.9740} & \textbf{0.9675} \\ 
        \hline
        \multirow{2}{*}{0.3} & Random & 0.9545 & 0.9220 & \textbf{0.9675} & 0.9545 & 0.9350 \\ 
        \cline{2-7}
        & USN-guided         & \textbf{0.9805} & \textbf{0.9350} & \textbf{0.9675} & \textbf{0.9610} & \textbf{0.9415} \\ 
        \hline
    \end{tabular}
\end{table}

\begin{figure}[htbp]
    \centering
    \begin{subfigure}[b]{0.45\linewidth}
        \centering
        \includegraphics[width=\linewidth]{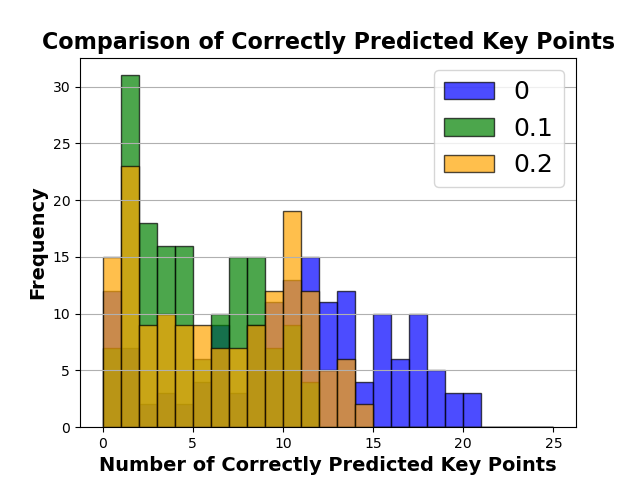}
        \caption{Correctly predicted keypoints (USN-guided pruning rates $\rho\in\{0, 0.1,0.2\}$)}
    \end{subfigure}
    \hfill
    \begin{subfigure}[b]{0.45\linewidth}
        \centering
        \includegraphics[width=\linewidth]{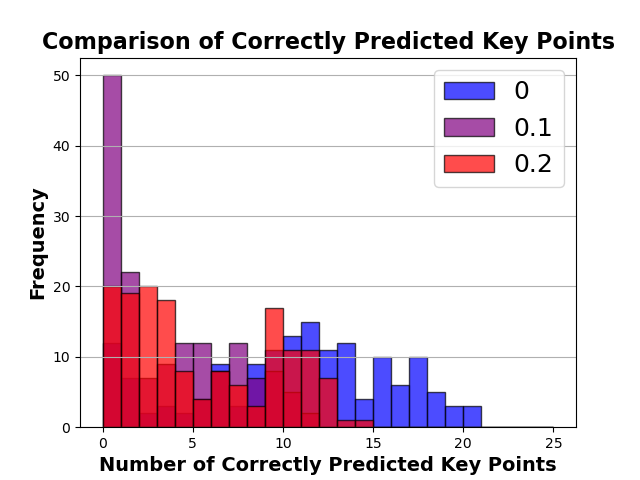}
        \caption{Correctly predicted keypoints (Random pruning rates $\rho\in\{0, 0.1,0.2\}$)}
    \end{subfigure}

    \vfill
    \begin{subfigure}[b]{0.45\linewidth}
        \centering
        \includegraphics[width=\linewidth]{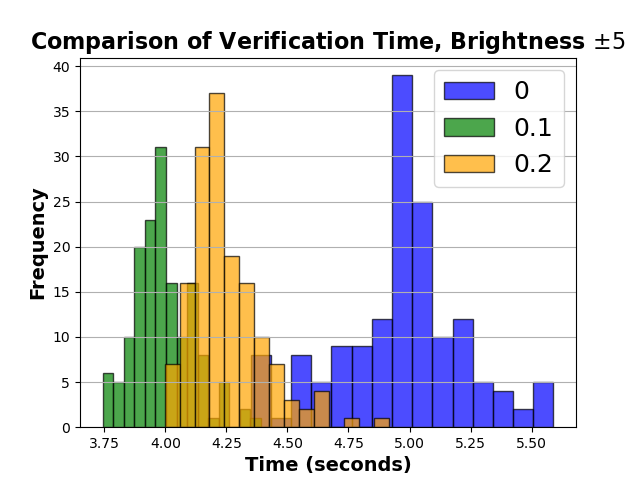}
        \caption{Average verification time ( USN-guided pruning $\rho\in\{0, 0.1,0.2\}$)}
    \end{subfigure}
    \hfill
    \begin{subfigure}[b]{0.45\linewidth}
        \centering
        \includegraphics[width=\linewidth]{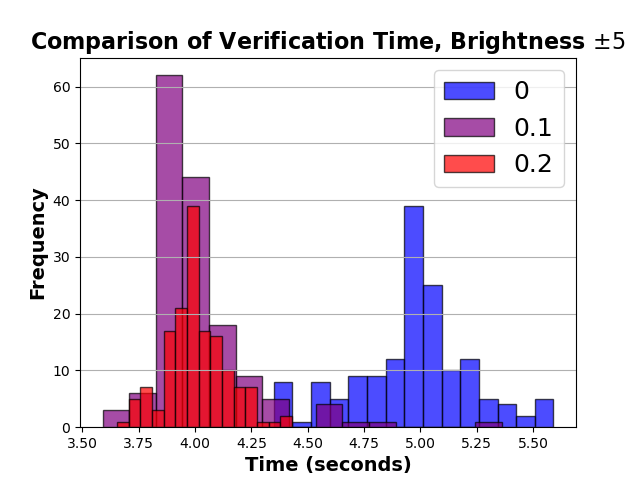}
        \caption{Average verification time ( Random pruning $\rho\in\{0, 0.1,0.2\}$)}
    \end{subfigure}

    \caption{Comparison of the number of correctly predicted keypoints and verification time under different pruning strategies for ResNet18 under Wasserstein regularization $\lambda_W=10$.}
    \label{fig:resnet18_pruning_comparison}
\end{figure}

\paragraph{ResNet18}
We compare pruning strategies in terms of fidelity (correct keypoints) and efficiency (verification time) in \Cref{fig:resnet18_pruning_comparison}. The model without pruning ($\rho=0$) results in more correct keypoints, but the verification time is significantly larger than that of the models with pruning. For pruned models under the same Wasserstein regularization $\lambda_W=10$,
compared to \emph{random pruning} baselines,  the \emph{USN-guided pruning} ones produce keypoint histograms that shift to the right in
\Cref{fig:resnet18_pruning_comparison} (a,b), indicating more correctly predicted keypoints; in
\Cref{fig:resnet18_pruning_comparison} (c,d), both achieve a comparable left-shift in runtime, indicating similar verification speedups.
In short, USN-guided pruning attains better efficiency with substantially higher fidelity than the random pruning baseline.
Among USN settings, the intermediate configuration offers the best efficiency-fidelity trade-off; within random pruning, the higher rate is less damaging than the lower rate yet remains inferior to USN-guided pruning.

\paragraph{Visualization of model weight after training} In \Cref{fig:visualization}, we visualize the fourth layer of the CNN7 model after three different training paradigms: vanilla training with only task losses without USN metrics, robust training with USN metrics but without pruning, and the proposed robust pruning with USN metrics.  It can be seen that the unbiased and smooth metrics for neurons after robust training with USN metrics but without pruning (\texttt{USN No Pruning}) are the highest due to over-parameterization by robust training \cite{wei2024improve}. However, ours (\texttt{USN Pruning}) can significantly reduce the number of neurons with higher values of unbiased and smooth metrics and retain most of the stable neurons, where our unbiased and smooth metrics  are comparable to those obtained after vanilla training. In summary, ours is easier to verify compared to the robust training of \texttt{USN No Pruning} and more robust than the model of vanilla training without USN.

\subsection{Ablation Study}
\label{sec:ablation}

\paragraph{Effect of pruning rate in \Cref{tab:pr_rate}}
Sweeping $\rho\in\{0.1,0.2,0.3\}$ indicates that $\rho=0.2$ offers the most balanced trade-off between model verifiability and expressiveness under input perturbations compared to the non-pruning baseline $\rho=0.0$. Due to the progressive pruning, the higher the pruning ratio is, the less expressive the pruned model will be. In contrast, a stronger pruning effect will make the pruned model easier to verify. At $\rho=0.1$, insufficient pruning fails to eliminate enough unstable neurons that contribute to over-approximation during verification, resulting in suboptimal certified robustness despite maintaining high model capacity (which is still lower than $\rho=0.0$ through). However, aggressive pruning $\rho=0.3$  removes too many essential neurons needed for robust feature representation, explaining the degradation at higher pruning ratios.

\begin{table}[htbp]
    \centering
    \caption{CNN7 verification accuracy under different pruning ratio under $\lambda_W=10$.}
    \label{tab:pr_rate}
    \setlength{\tabcolsep}{6pt}   
    \renewcommand{\arraystretch}{1.30} 
    \begin{tabular}{|c|c|c|c|c|c|}
        \hline
        \multirow{2}{*}{Pruning Ratio} & \multicolumn{2}{c|}{Brightness} & \multicolumn{3}{c|}{Contrast} \\ 
        \cline{2-6}
        & $\pm2$ & $\pm5$ & $\pm0.01$ & $\pm0.02$ & $\pm0.05$ \\ 
        \hline
        $\rho=0.0$& 0.9805 & \textbf{0.9480} & 0.9870 & \textbf{0.9870} & 0.9545 \\ 
        \hline
        $\rho=0.1$ & 0.9740 & 0.9285 & 0.9805 & 0.9545 & 0.8961 \\ 
        \hline
        $\rho=0.2$ & \textbf{0.9870} & \textbf{0.9480} & \textbf{0.9935} & 0.9740 & \textbf{0.9675} \\ 
        \hline
        $\rho=0.3$ & 0.9805 & 0.9350 & 0.9675 & 0.9610 & 0.9415 \\ 
        \hline
    \end{tabular}
\end{table}

\paragraph{Effect of Wasserstein Regularization in \Cref{tab:W_weight}}
Enabling the Wasserstein term ($\lambda_{W}=10$) generally improves certification at $\rho\in\{0.2,0.3\}$, with observable gains across all tested perturbations at the moderate rate and the largest improvements at higher contrast and stronger brightness at the stronger rate. At $\rho=0.1$, however, the Wasserstein regularizer can over-concentrate pruning and degrade robustness under strong perturbations, indicating that the regularization term is most beneficial under mild perturbations with satisfactory accuracy.

\begin{table}[htbp]
    \centering
    \caption{CNN7 verification accuracy with or without Wasserstein regularization ($\lambda_W\in\{10,0\}$) under different pruning ratio $\rho$.}
    \label{tab:W_weight}
    \renewcommand{\arraystretch}{1.3}
    \resizebox{0.96\columnwidth}{!}{%
    \begin{tabular}{|c|c|c|c|c|c|c|}
        \hline
        \multirow{2}{*}{$\rho$} & \multirow{2}{*}{Wasserstein weight} &
        \multicolumn{2}{c|}{Brightness} & \multicolumn{3}{c|}{Contrast} \\
        \cline{3-7}
        & & $\pm2$ & $\pm5$ & $\pm0.01$ & $\pm0.02$ & $\pm0.05$ \\
        \hline
        \multirow{2}{*}{0.1} & $\lambda_W=0$  & 0.9610 & \textbf{0.9480} & 0.9740 & \textbf{0.9610} & \textbf{0.9415} \\
        \cline{2-7}
                              & $\lambda_W=10$ & \textbf{0.9740} & 0.9285 & \textbf{0.9805} & 0.9545 & 0.8961 \\
        \hline
        \multirow{2}{*}{0.2} & $\lambda_W=0$  & 0.9675 & 0.9350 & 0.9610 & 0.9675 & 0.9350 \\
        \cline{2-7}
                              & $\lambda_W=10$ & \textbf{0.9870} & \textbf{0.9480} & \textbf{0.9935} & \textbf{0.9740} & \textbf{0.9675} \\
        \hline
        \multirow{2}{*}{0.3} & $\lambda_W=0$  & 0.9545 & 0.9090 & \textbf{0.9740} & 0.9415 & 0.8311 \\
        \cline{2-7}
                              & $\lambda_W=10$ & \textbf{0.9805} & \textbf{0.9350} & 0.9675 & \textbf{0.9610} & \textbf{0.9415} \\
        \hline
    \end{tabular}%
    }
\end{table}

\section{CONCLUSION}

This paper addresses over-parameterization challenges in robustness certification by introducing Unbiased and Smooth Neuron (USN) metrics that identify neurons contributing to certification looseness under semantic perturbations. Our progressive training pipeline integrates USN-guided pruning with Wasserstein distance regularization to achieve structured sparsity while preserving model expressiveness. Experiments on keypoint detection under brightness and contrast variations demonstrate that strategic removal of high-variance neurons consistently improves both certification accuracy and verification efficiency compared to unpruned and randomly pruned baselines. This work establishes a principled foundation for robustness-aware model compression, enabling more tractable formal guarantees for safety-critical applications while maintaining task performance.

\section*{ACKNOWLEDGMENT}
This material is based upon work supported by The Boeing Company. Any opinions, findings, and conclusions
or recommendations expressed in this material are those of the authors and do not necessarily reflect the
views of The Boeing Company.
\bibliographystyle{abbrv}   

\bibliography{root}

\end{document}